\documentclass[journal]{IEEEtran}
\usepackage{amsmath,graphicx}
\usepackage{color}
\usepackage{url}
 \usepackage{setspace}

\newcommand{\green}[1]{{\color[rgb]{0,0,0}#1}}

\usepackage{multirow}
\usepackage{adjustbox}

\title{Fast Mesh Data Augmentation via Chebyshev Polynomial of Spectral filtering}

\author{Shih-Gu Huang, %~\IEEEmembership{Member,~IEEE,}
       Moo K. Chung,
         Anqi Qiu,
        and Alzheimer's Disease Neuroimaging Initiative% <-this % stops a space
\thanks{S.-G. Huang is with Department of Biomedical Engineering, National University of Singapore, Singapore (e-mail: shihgu@gmail.com).}% <-this % stops a space
\thanks{M.K. Chung is with the Department of Biostatistics and Medical Informatics, University of Wisconsin, Madison, WI 53706, USA (e-mail: mkchung@wisc.edu).}
\thanks{A. Qiu is with Department of Biomedical Engineering and the N.1 Institute for Health, National University of Singapore, Singapore, 117583 (e-mail: bieqa@nus.edu.sg).}
}

\begin{document}
\maketitle
%\IEEEpeerreviewmaketitle

%\thispagestyle{plain}   % Show page number. Remove it later
%\pagestyle{plain}        % Show page number. Remove it later

\begin{abstract}
Deep neural networks have recently been recognized as one of the powerful \green{learning} techniques in  computer vision and medical image analysis. Trained deep neural networks need to be generalizable to \green{new data that was not seen before}. In practice, there is often insufficient training data available and augmentation is used to expand the dataset. Even though graph convolutional neural network (graph-CNN) has been widely used in deep learning, there is a lack of augmentation methods to generate data on graphs or surfaces. This study proposes two unbiased  augmentation methods,  Laplace-Beltrami eigenfunction Data Augmentation (LB-eigDA) and Chebyshev polynomial Data Augmentation (C-pDA), to generate new data on surfaces, whose mean is the same as that of real data. LB-eigDA augments data via the resampling of the LB coefficients. In parallel with LB-eigDA, we introduce a fast augmentation approach, C-pDA, that employs a polynomial approximation of LB spectral filters on surfaces. We design LB spectral bandpass filters by Chebyshev polynomial approximation and resample signals filtered via these filters to generate new data on surfaces. We first validate LB-eigDA and C-pDA via simulated data and demonstrate their use for improving classification accuracy. We then employ the brain images of Alzheimer's Disease Neuroimaging Initiative (ADNI) and extract cortical thickness that is represented on the cortical surface to illustrate the use of the two augmentation methods. We demonstrate that augmented cortical thickness has a similar pattern to real data. Second, we show that C-pDA is much faster than LB-eigDA. Last, we show that C-pDA can improve the AD classification accuracy of graph-CNN. 
\end{abstract}

%\begin{keywords}
%Data augmentation, signals on surfaces, Laplace-Beltrami operator, cortical thickness
%\end{keywords}

\section{Introduction}

Deep neural networks have recently been recognized as one of the powerful \green{learning} techniques in  computer vision and medical image analysis \cite{litjens2017survey,shen2017deep}.  Training deep neural networks requires a large dataset so that they are generalizable to data that have never been seen before. This is challenging especially in the field of medical image analysis. Building big medical image datasets is expensive and labor-intensive to collect, and is related to patient privacy, and the requirement of medical experts for labeling. Not having enough data could overfit training data so that network models are not generalized to new data. Moreover, studies on rare diseases or medical screening also face the problem of class imbalance \green{with} a skewed ratio of majority to minority samples  \cite{mazurowski2008training,ker2017deep}. 
These obstacles have led to many studies on image data augmentation (see review in \cite{leevy2018survey}).  Data augmentation assumes that additional information can be extracted from an original dataset. It is a very powerful \green{approach for overcoming overfitting in deep learning.}

Image augmentation inflates the size of training data via either image transformation or oversampling. New images can be generated by warping  existing images via geometric (rotation, flipping) and color transformations \cite{shorten2019survey}, random erasing \cite{zhong2017random}, and adversarial training \cite{goodfellow2014explaining,ganin2016domain} such that their labels are preserved. In contrast, oversampling augmentation creates synthetic data by mixing existing images, auto encoder-decoder \cite{kingma2013auto,devries2017dataset}, and generative adversarial networks (GANs)  \cite{goodfellow2014generative,yi2019generative}. Even though GANs are powerful, their computation is more expensive compared to image warping methods.

Among existing image augmentation methods  \cite{shorten2019survey,zhong2017random,kingma2013auto,devries2017dataset,goodfellow2014generative,yi2019generative}, image data are defined on an equi-spaced grid in the Euclidean space. However, medical images in the Euclidean space may not fully characterize the geometry of human organs that encompass their intrinsic and complex anatomy, as well as physiological functions. For example, the cerebral cortex is composed of ridges (gyri) and valleys (sulci). Due to the way  gyri and sulci are curved, the cortex is thicker in gyri but thinner in sulci. Hence, it is preferred to represent brain images in a way that the underlying geometrical information is encoded. One can express the cerebral cortex as a surface embedded in the 3D Euclidean space. Existing literature has demonstrated that such representation incorporates useful geometry information of the brain into machine learning for disease diagnosis \cite{YangQiu2013,davatzikos_biopsy_2008,apostolova_brain_2006,anqi_adni_2009}. Recently, a number of deep neural networks, such as diffusion-convolutional neural networks (DCNNs) \cite{atwood2015diffusion}, PATCHY-SAN \cite{niepert2016learning,duvenaud2015conv}, gated graph sequential neural networks \cite{yujia2015gated}, DeepWalk \cite{perozzi2014deepwalk}, and spectral graph convolutional neural networks (graph-CNN)   \cite{bruna2013spectral,Defferrard2016,henaff2015deep,kipf2016semi,liy2016syncspeccnn,ktena2017distance,shuman2016vertex} can take data on surfaces for classification. The core challenge for implementing CNN on surfaces lies in defining the convolution on surfaces. These existing neural network approaches focus on  how to process vertices whose neighborhood has different sizes and connections for the convolution in the spatial domain. Alternately, the convolution can be defined as a multiplication involving a diagonal matrix in \green{the} graph Fourier transform derived from a normalized graph Laplacian in the spectral domain. Hence, existing image warping augmentations on equi-spaced grids (e.g., flipping, rotation, shifting) may not directly apply to data on surfaces since the points on surfaces are not on the equi-spaced grid of the Euclidean space. Nevertheless, there is a lack of augmentation approaches to generate data on surfaces.

This study proposes two unbiased augmentation methods,  Laplace-Beltrami eigenfunction Data Augmentation (LB-eigDA) and Chebyshev polynomial Data Augmentation (C-pDA), to generate new data on surfaces. These two approaches preserve the mean of real data in each class, which is crucial for classification problems. These two approaches are motivated by the Fourier representation of signals in equi-spaced Euclidean grids. A signal in equi-spaced Euclidean grids can be created as a linear combination of Fourier bases, where the corresponding Fourier coefficients can be generated via the resampling of the Fourier coefficients of existing signals \cite{tang.2013,ravanbakhsh.2016,wang.2018.annals}. We adopt this idea and compute the eigenfunctions of the Laplace-Beltrami (LB) operator on a surface. New data on the surface can be constructed via the resampling of the LB coefficients among real data on the surface. 
%\footnote{If I remember I read "preservation of mean of original data" 3-4 times by now. But no idea why that matters or if it is an even important issue. Don't you think this is too much repetition without providing some hint on the importance? Need more tight editing.}
%Even though the LB-eigDA is straightforward, its computation is expensive if a surface is large \cite{huang.2019.TMI}.

In parallel with LB-eigDA, we introduce a fast augmentation approach, C-pDA, that employs a polynomial approximation of LB spectral filters on surfaces. C-pDA is designed to be in line with graph-CNN \cite{Defferrard2016,shuman2016vertex}, where spectral filters are implemented via Chebychev  polynomial approximation such that the resulting convolution can be written  as a polynomial of the adjacency matrix of a graph. This  avoids the cost of calculating the eigenfunctions of a large-scale graph Laplacian. In \cite{Defferrard2016,shuman2016vertex}, it is shown that the $k$-th order Chebyshev polynomial formation of the graph Laplacian is equivalent to $k$-ring filtering. In C-pDA, we design LB spectral bandpass filters by Chebyshev polynomial approximation and resample filtered real data to generate new data. Due to the recurrence relation of Chebyshev polynomials, the computation of the C-pDA method can be efficient. We validate LB-eigDA and C-pDA  using simulated data with the ground truth of class labels. We further employ the methods to the cortical surface data in Alzheimer's Disease Neuroimaging Initiative (ADNI). We first demonstrate that augmented cortical thickness data have a similar pattern to real data. Second, we show that C-pDA is much faster than LB-eigDA . Last, we illustrate the use of C-pDA to improve the AD classification of the graph-CNN  \cite{Defferrard2016}.

The main contributions of this study are as follows. 
\begin{itemize}
\item We introduce two augmentation methods to generate new data on surfaces using the LB eigenfunctions and LB spectral filters.
\item We show that C-pDA is computationally more efficient than LB-eigDA.
\item We demonstrate that C-pDA improves the graph-CNN performance on the classification of AD patients.
\end{itemize}

\section{Methods}

\subsection{Augmentation based on the Laplace-Beltrami representation of signals on a surface mesh}
\label{sec:lbeig}
We introduce a data augmentation method based on the Laplace-Beltrami representation of signals on a surface mesh. 
%\footnote{Still I have no idea why you are trying to preserve mean value...!!! Explain it in introduction early!!!!}
We denote the surface as $\mathcal{M}$ with the Laplace-Beltrami (LB) operator $\Delta$ on $\mathcal{M}$. Let $\psi_j$ be the $j^{th}$ eigenfunction of the LB-operator with eigenvalue $\lambda_j$
\begin{equation}\label{eq:eigs}
{\Delta}\psi_j=\lambda_j\psi_j \ ,
\end{equation}
where $0=\lambda_0 \leq \lambda_1 \leq \lambda_2 \leq \cdots$. A signal $f(x)$ on the surface $\mathcal{M}$ can be represented as a linear combination of the LB eigenfunctions
\begin{equation}\label{eq:x}
f(x)=\sum_{j=0}^{\infty}c_j\psi_j(x) \  ,
\end{equation}
where $c_j$ is the $j^{th}$ coefficient associated with the eigenfunction $\psi_j(x)$. For $n$ observations, $f_1(x), \cdots, f_n(x)$,  $f_i(x)$ can be represented as 
$$
f_i(x) = \sum_{j=0}^{\infty}c_j^{(i)} \psi_j(x) \  ,
$$
where $c_j^{(i)}$ is the $j^{th}$ coefficient associated with the $j^{th}$ LB eigenfunction for the $i^{th}$ observation. We like to generate new data 
%such that the mean value of the new data is the same as that of the real data 
based on the frequency resampling of these $n$ observations. This is similar to creating new samples via permuting Fourier coefficients  \cite{wang.2018.annals}. 
%\footnote{When we will find out why we have to sample with respect to mean?}  
Let $S_n$ be the permutation group of order $n$ \cite{chung.2019.CNI} and $\tau \in S_n$ be an element of permutation given by
\begin{equation}
\label{eqn:pi}
\tau=\left(  \begin{array}{cccc}
1 & 2 & \cdots & n\\
\tau(1) & \tau(2) & \cdots & \tau(n)
\end{array}  \right).
\end{equation}
$\tau(i)$ indicates element $i$ is permuted to $\tau(i)$. 
We resample the LB coefficients to 
obtain new data representation $f_{i^\prime}(x)$:
\begin{equation}
\label{eq:aug_eig}
f_{i^\prime}(x) =  \sum_{j=0}^{\infty}c_j^{\tau_j(i)}  \psi_j(x) \  ,
\end{equation}
where $\tau_j(\cdot)$ is %a resampling function  
 the permutation
on the $j^{th}$ LB coefficients among the $n$ observations. We will refer this approach as {\it{LB eigenfunction Data Augmentation (LB-eigDA)}}.

Based on Eq. (\ref{eq:aug_eig}), one can show that the mean of $f_{i^\prime}(x)$ \green{over every possible permutation is} the same as that of $f_{i}(x)$ since the permutation function $\tau(\cdot)$ does not change the mean of the LB coefficients. 

\subsection{Augmentation via Chebyshev polynomials }
Previous research suggests that the augmentation strategy of Gaussian filters leads to the best validation accuracy in medical imaging classification tasks \cite{hussian_AMIA_2017}.  We now introduce the second data augmentation approach, {\it{Chebyshev polynomial Data Augmentation (C-pDA)}}. The idea of C-pDA is similar to the augmentation strategy of Gaussian filters in equi-spaced grids of the Euclidean space by designing LB spectral filters on surfaces. We design LB spectral filters that are similar to spectral filter banks \cite{tan.2015}. We can then approximate real data on surfaces using these LB spectral filters and resample the LB spectral filtered signals of real data in order to generate new data on surfaces. To avoid the direct computation of the LB eigenfunctions, we will employ the Chebyshev polynomial approximation of LB spectral filters, which is computationally efficient.   In the following, we first describe the Chebyshev polynomial approximation of an LB spectral filter and then design LB spectral bandpass filters for the C-pDA approach.

\subsubsection{Chebychev polynomial approximation of LB spectral filters}
Consider an LB spectral filter $g$ on the surface $\mathcal{M}$ with spectrum $g(\lambda)$ as
\begin{equation}\label{eq:g}
g(x,y)=\sum_{j=0}^{\infty}g(\lambda_j)\psi_j(x)\psi_j(y).
\end{equation}
Based on Eq. (\ref{eq:x}), the convolution of a signal $f$ with the filter $g$ can be written as  
\begin{equation}\label{eq:gx}
 h(x) = g \ast f(x)
=\sum_{j=0}^{\infty}g(\lambda_j) c_j\psi_j(x).
\end{equation}
As suggested in \cite{Defferrard2016,wee2019cortical,coifman2006diffusion,Hammond2011129,kim2012wavelet,tan.2015}, the filter spectrum $g(\lambda)$ in  Eq. (\ref{eq:gx}) can be represented as the expansion of Chebyshev polynomials, $T_k, k=0, 1, 2, \dots, \infty$, such that
\begin{equation}
\label{eq:glambda}
g(\lambda)=\sum_{k=0}^{\infty}\theta_k T_k(\lambda) \ .
\end{equation}
$\theta_k$ is the $k ^{th}$ expansion coefficient associated with the $k ^{th}$ Chebyshev polynomial.  $T_k$ is the Chebyshev polynomial of the form
$T_k(\lambda)=\cos(k\cos^{-1}\lambda)$ with recurrence
$$T_{k+1}(\lambda)= (2-\delta_{k0})\lambda \ T_{k}(\lambda)- T_{k-1}(\lambda),$$
where $\delta_{k0}$ is Kronecker delta.  %such that $\delta_{k0}=1$ when $k=0$ and 0 otherwise. 
The convolution in Eq. (\ref{eq:gx}) can be rewritten as 
\begin{equation}\label{eq:gx3}
h(x) = g \ast f(x)=\sum_{k=0}^{\infty}\theta_k T_k(\Delta) f(x).
\end{equation}
This Chebyshev polynomial approximation of the spectral filter has previously used in diffusion wavelet transform  \cite{Hammond2011129,coifman2006diffusion,kim2012wavelet,donnat2018learning},  graph convolutional neural network \cite{Defferrard2016,wee2019cortical}, spectral wavelet  transform  \cite{tan.2015}, and heat diffusion \cite{huang2020fast} on graphs. The polynomial method avoids the direct computation of the LB eigenfunctions through the recursive computation of $T_k(\Delta)f(x)$ and preserves local geometric structure of the surface \cite{Defferrard2016,huang2020fast}.

\subsubsection{C-pDA}
%\footnote{With suddenly new notations, this section is not connecting well from previous section. Also how did you get equation (8)? Some explanation needed.}
We design a series of LB spectral bandpass filters, $g_l,\ l=1, 2, \dots, L$, based on Eq. (\ref{eq:glambda}) such that
$$
g_l(\lambda)=\sum_{k=0}^{\infty}\theta_{lk}T_k(\lambda) \,
$$
where $\theta_{lk}$ is the $k ^{th}$ Chebyshev expansion coefficient of the $l ^{th}$ bandpass filter. The frequency band of the $l ^{th}$ bandpass filter is $\lambda \in [\epsilon_l \ \epsilon_{l+1}]$. Now, a signal $f(x)$ on surface $\mathcal{M}$ can be approximated using these filters such that 
\begin{equation}
\label{eq:xfir}
f(x) \approx h_0+\sum_{l=1}^L g_l(\Delta) f(x),
\end{equation} 
where $h_0$ is the mean of $f(x)$ over the surface.  If $g_l,\  l=1,2,\dots, L,$ together span the entire spectrum of $f(x)$, then the spectral information of $f(x)$ is retained.

%\footnote{Red part is written in a convoluted fashion. Need better rewriting.}
%\footnote{ \red{Where that resampling that preserves mean has gone? The paper talked it about 5 times and the issue completely gone without any sort of explanation?}} 
%Assume that the spectral domains of \blue{$g_l$} are not overlapped. 
%\footnote{What does previous sentence has to do with the next sentence.} 

We develop the C-pDA approach in a way similar to the LB-eigDA approach in Eq. (\ref{eq:aug_eig}) such that
\begin{equation}
\label{eq:aug_cp}
f_{i^\prime}(x) =  h_0^{\tau_0(i)} + \sum_{l=1}^{L}\Big(g_l(\Delta) f_{i}(x) \Big)^{\tau_l(i)} \  ,
%= h_0^{\tau_0(i)} + \sum_{l=1}^{L}g_l(\Delta) f_{\tau_l(i)}(x)  \  ,
\end{equation}
where $\tau_l(\cdot)$ is the permutation on the $l^{th}$ filtered signal among the $n$ observations $f_1$, $f_2$,..., $f_n$ such that the $i^{th}$ observation is permuted to the $\tau_l(i)^{th}$ observation. Hence, C-pDA generates new data via resampling the $l^{th}$ filtered outputs among the $n$ observations and summing the resampled signals across $L$ filters. Again, we can show that the mean of $f_{i^\prime}(x)$ \green{over every possible permutation} is the same as that of $f_{i}(x)$ since the permutation function $\tau(\cdot)$ does not change the mean of the filtered signals.

With the Chebyshev polynomial approximation, we can rewrite Eq. (\ref{eq:aug_cp}) as 
\begin{equation}
\label{eq:aug_cpR}
f_{i^\prime}(x) =  h_0^{\tau_0(i)} +\sum_{l=1}^{L}  \Big(\sum_{k=0}^{\infty} \theta_{lk} T_k(\Delta) f_i(x) \Big)^{\tau_l(i)}\  .
\end{equation}

\begin{figure}[t]
\centering
\includegraphics[width=1\linewidth,clip=true]{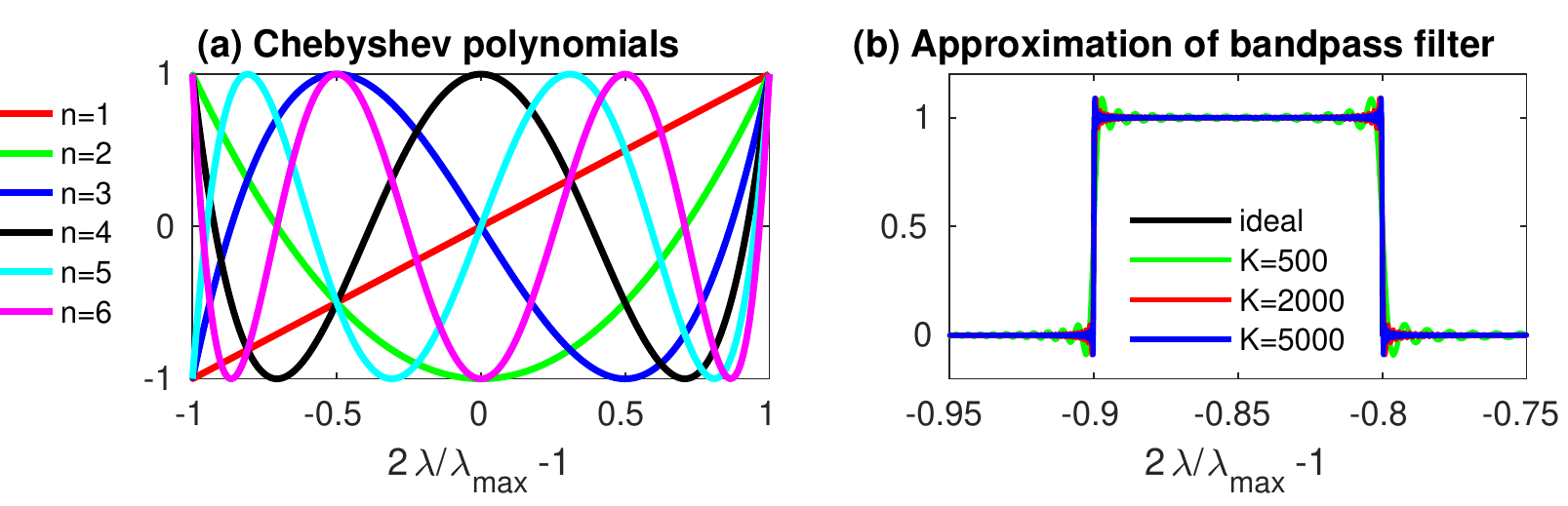}
\caption{
(a) Chebyshev polynomials of order 1 to 6. (b) An ideal rectangular bandpass filter with  range $\lambda\in[0.05\lambda_{\max},0.1\lambda_{\max}]$ and its approximation of Chebyshev polynomials of order up to $K=500, 2000$, and $5000$.
}
\label{fig:BW}
\end{figure}

%When a set of spectral filters cover the whole spectrum of signals, the filter outputs can also be used to represent the signals. The representation by LB eigenfunctions in Eq. \eqref{eq:x} can be deemed as a special case of the filter representation when each filter is an impulse function covering only one eigenvalue. Then, we can apply \blue{a} similar resampling strategy to the filter representation \blue{like LB-eigDA. Moreover,} using spectral filters has the advantage of implementation by fast polynomial approximation, which also benefits the computation of the C-pDA method.} \footnote{We do need red part. I read like 5 times and still no idea what it even means. It talks in very ambiguous unspecific way such that it fail to deliver message. Can you write far better or delete it if not needed.}

\subsection{LB-eigDA and C-pDA numerical implementation }
For the implementation of the LB-eigDA in Eq. (\ref{eq:aug_eig}), we adopt the discretization scheme of the LB operator in \cite{tan.2015}, where surface $\mathcal{M}$ is represented by a triangulated mesh with a set of triangles and vertices $v_i$. The $ij^{th}$ element of the LB-operator on $\mathcal{M}$ can be computed as 
\begin{equation}
\label{eq:Delta}
\Delta_{ij}= C_{ij} /A_i,
\end{equation}
where $A_i$ is the Voronoi area of vertex $v_i$ if the triangles containing $v_i$ are nonobtuse \cite{meyer2003discrete} and Heron's area if the triangles containing $v_i$ are obtuse \cite{tan.2015,meyer2003discrete}. The off-diagonal entries are defined as $C_{ij}=-(\cot\theta_{ij}+\cot\phi_{ij})/2$ if $v_i$ and $v_j$ form an edge, otherwise $C_{ij}=0$. The diagonal entries $C_{ii}$  are computed as $C_{ii}=-\sum_{j} C_{ij}$. Other cotan discretizations of the LB operator are discussed in \cite{chung.2004.ISBI,qiu.2006, chung.2015.MIA}. When the number of vertices on $\mathcal{M}$ is large, the computation of the LB eigenfunctions can be costly \cite{huang.2019.TMI}.

For the numerical implementation of the C-pDA method in Eq. (\ref{eq:aug_cpR}), we need to first determine the order of Chebyshev polynomials while $g_l(\lambda)$ have less overlap for C-pDA. One can quantify the overlap among the filters $g_l$ via training the spectral band between the passband and stopband \cite{oppenheim2001discrete}. A higher-order filter has a narrower transition band than a lower-order filter. Fig.~\ref{fig:BW} shows the transition bandwidth over order $K$ for Chebyshev polynomials when the filter band is $\lambda\in[0.05\lambda_{\max},0.1\lambda_{\max}]$, where $\lambda_{\max}$ is the maximum eigenvalue of the LB operator. In this study,  we empirically determined the order of Chebyshev polynomials as $K=5000$ for C-pDA, which achieves the transition bandwidth as small as $3.5\times10^{-4}$ as illustrated in Fig.~\ref{fig:BW}. $L$ depends on the spectral distribution of the observations and thus is application specific. This study empirically determines $L$ in the below applications. 
%\footnote{How? Need explanation.} 

We take the advantage of the recurrence relation of the Chebyshev polynomials and compute C-pDA recursively. We now describe steps for the numerical implementation of Eq. (\ref{eq:aug_cpR}).
\begin{itemize}
\item[1.] discretize the surface  $\mathcal{M}$ using a triangulated mesh;  
  \item[2.] compute $\Delta$ based on Eq. (\ref{eq:Delta}) for the surface mesh $\mathcal{M}$;
  \item[3.] compute the maximum eigenvalue $\lambda_{\max}$ of $\Delta$. For the standardization across surface meshes, we normalize $\Delta$ as $\widetilde\Delta=\frac{2\Delta}{\lambda_{\max}}-I$, where $I$ is an identity matrix;
 \item[4.] for the signal $f_i$ of the $i^{th}$ subject, compute  $T_k(\widetilde\Delta)f_i(x)$ recursively by
$$
T_{k+1}(\widetilde\Delta) f_i(x)= (2-\delta_{k0})\widetilde\Delta \ T_{k}(\widetilde\Delta) f_i(x)- T_{k-1}(\widetilde\Delta) f_i(x)
$$ 
with initial conditions 
$$T_{-1}(\widetilde\Delta)f_i(x)=0$$
 and  
$$T_0({\widetilde\Delta})f_i(x)=f_i(x).$$ 
%\footnote{\red{ is it the mean of $f_i(x)$ on the right hand? If so, we need to modify. It's still unclear why the resampling is mean preserving.}}
 \item[5.] compute each augmented signal $f_i'$ recursively as
 $$
f_{i^\prime}^k(x) = f_{i^\prime}^{k-1}(x) +\sum_{l=0}^{L}   \Big( \theta_{lk} T_k(\widetilde\Delta) f_i(x) \Big)^{\tau_l(i)}\  , 
$$
\end{itemize}
where $$
\theta_{lk} =   \frac{2-\delta_{k0}}{\pi} \int_{\epsilon_k}^{\epsilon_{k+1}}  T_k(\lambda) \frac{d\lambda}{\sqrt{1-\lambda^2}} \ ,
$$
where $[\epsilon_k \ \epsilon_{k+1}]$ is the frequency band of $g_l$. 
Steps 4 and 5 are repeated from $k=0$ till $k=K-1$. In step 5,  there is no need to explicitly compute each filtered signal, which saves computational time and memory, especially when a large number of filters are used.

\section{Simulation Experiments}

A majority of medical applications often face two challenges, limited sample sizes and potential uncertainty of diagnosis \cite{ranginwala2008clinical,tong2014multiple}. We designed simulation experiments with the ground truth of group labels to illustrate the use of LB-eigDA and C-pDA in the sample size estimation and diagnosis classification. 

\begin{figure}[t]
\centering
\includegraphics[width=1\linewidth]{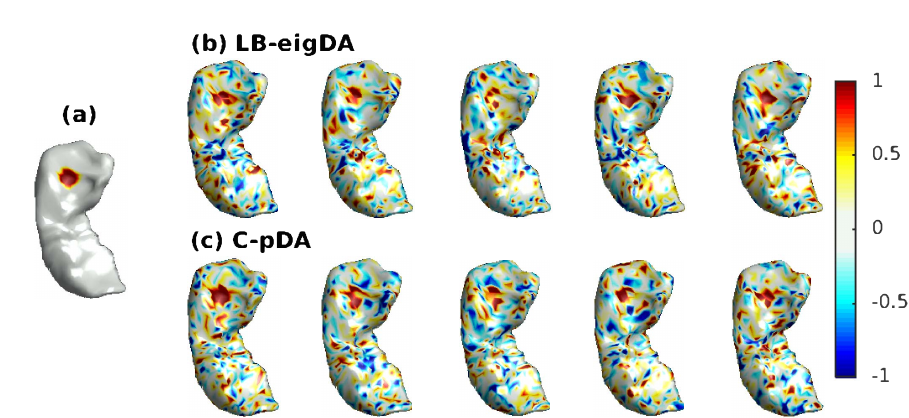}
\caption{Simulated and augmented data in Group 1. (a) Averaged signal over 500 data that were simulated via the distribution  ${\cal N}(1, \sigma^2)$ in the small patch (red region) of the hippocampus and the distribution of ${\cal N}(0, \sigma^2)$ at each vertex on the rest of the hippocampus. (b) five augmented data for Group 1 via the LB-eigDA method ; and (c) five augmented data for Group 1 via the C-pDA method. }
\label{fig:simulation1}
\end{figure}

We performed simulation experiments using a hippocampus surface mesh with 1184 vertices and 2364 triangles. We generated two groups of simulated data on this surface mesh: $n$ samples in Group 0 and $m$ samples in Group 1. We first generated $n+m$ measurements by a normal distribution with mean $0$ and variance $\sigma^2$, i.e., ${\cal N}(0, \sigma^2)$, at each vertex of the hippocampus surface. The first $n$ measurements were considered as samples in Group 0, while the rest of $m$ measurements were added signal $1$ in a small patch on the hippocampus (see the red region in Fig. ~\ref{fig:simulation1} (a)) and were considered as Group 1. Thus, Group 0 had the distribution  ${\cal N}(0, \sigma^2)$ at each vertex, while Group 1 had the distribution  ${\cal N}(1, \sigma^2)$ in the small patch of the hippocampus and the distribution of ${\cal N}(0, \sigma^2)$ at each vertex on the rest of the hippocampus. Fig.~\ref{fig:simulation1} (a) shows the signal averaged over 500 samples in Group 1.

To generate augmented data, we computed all the 1184 eigenfunctions for LB-eigDA. The hippocampal surface mesh had the spectrum over $[0,\ 10.9]$. For C-pDA,  we used 109 bandpass filters whose bandwidth was 0.1 and a mean filter that computed the average value of a signal over the hippocampal surface. Each filter was approximated by Chebyshev polynomials of order 5000. Fig.~\ref{fig:simulation1}(b) and (c) show 5 augmented data generated by LB-eigDA and C-pDA for Group 1, respectively.

\begin{figure}[t]
\centering
\includegraphics[width=1\linewidth]{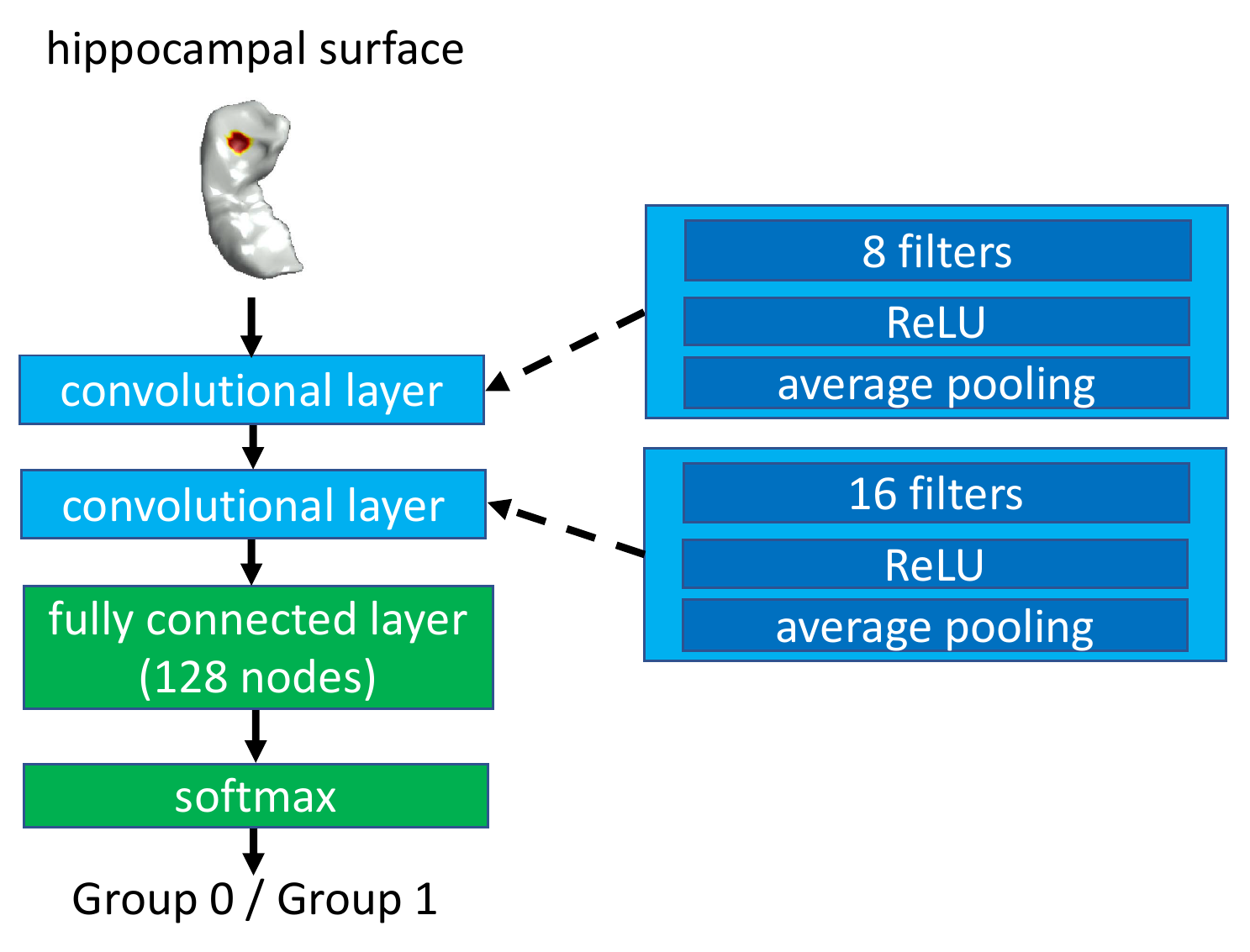}
\caption{The LB-based spectral CNN with 2 convolutional layers and one fully connected layer. Each convolutional layer is comprised of filters approximated by the Chebyshev polynomials of order $7$, a rectified linear unit (ReLU), and average pooling. }
\label{fig:architecture}
\end{figure}

\begin{figure}[t]
\centering
\includegraphics[width=1\linewidth]{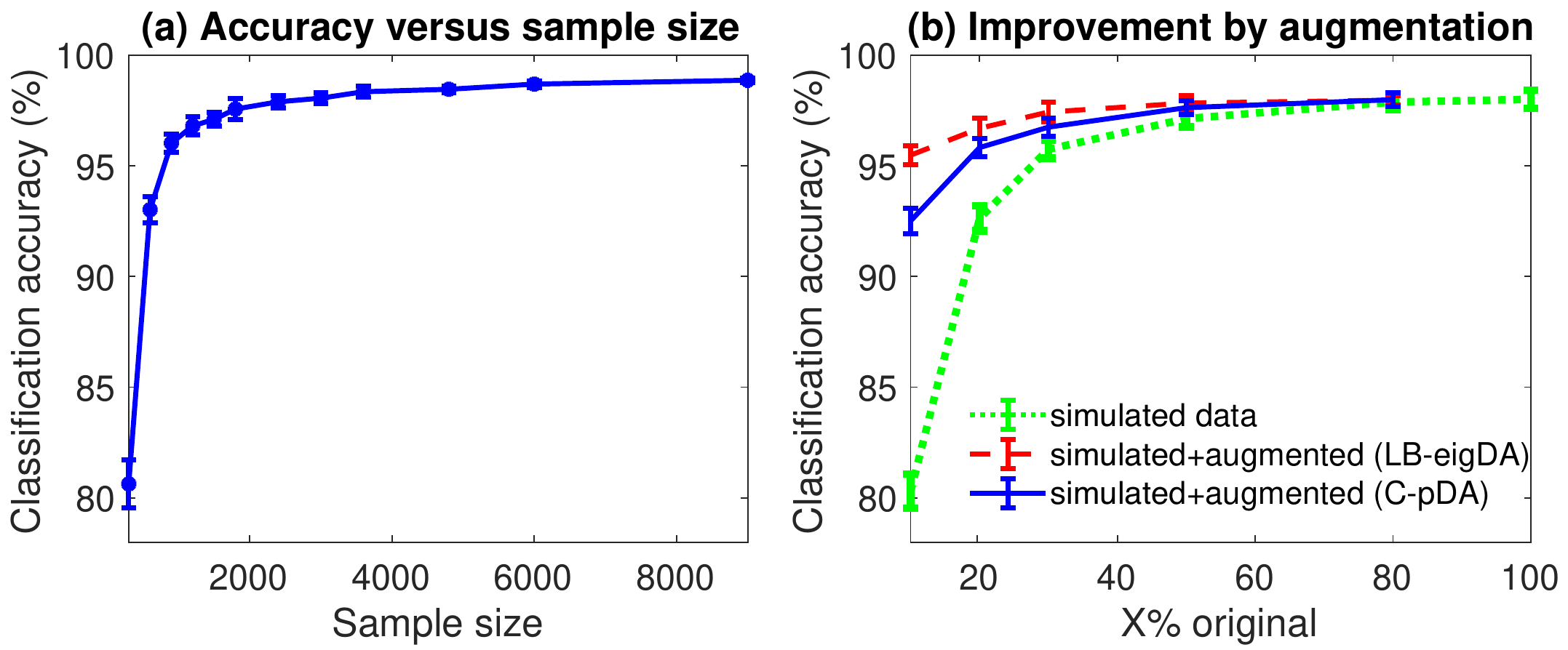}
\caption{Classification accuracy on simulated and augmented data. (a) The classifications accuracy using simulated data  when the sample size increase from 500 to 9000. (b) The green dotted line shows the classification accuracy when only $X\%$ of simulated data was used as the training set. The red dashed and blue solid lines show the classification accuracy when only $X\%$ of the training set
were simulated data and $1-X\%$ of the training set were augmented data by the LB-eigDA and C-pDA methods, respectively. }
\label{fig:simulation2}
\end{figure}

We employed a convolutional neural network (CNN) that was a modified version of the graph-CNN in  \cite{Defferrard2016,wee2019cortical}. We employed the LB operator instead of the graph Laplacian in the CNN in this study. We called it as an LB-based spectral CNN. Fig.~\ref{fig:architecture}) shows the LB-based spectral CNN architecture with two convolutional layers due to the relatively small surface mesh of the hippocampus and one fully connected layer. The two convolutional layers had 8 and 16 filters, respectively. Each filter was characterized by the Chebyshev polynomials of order 7. Moreover, each layer also included a rectified linear unit (ReLU) and average pooling. We trained the network with an initial learning rate of $10^{-3}$, and a learning rate decay of $0.05$ for every $20$ epochs.  We applied the ten-fold cross-validation, where one fold was used for testing and the other 9 folds were for training ($75\%$) and validation ($25\%$).  Fig.~\ref{fig:simulation2}  (a) shows the classification accuracy versus total sample size $n+m$ with ratio $n/m=2$, which was similar to real ADNI data used below in this study. $\sigma=0.6$ was used. A higher value of $\sigma$ resulted in a similar curve except that more samples were required to reach the same classification accuracy. The accuracy reached  $98.1\%$ when the total sample size was $3000$ and then increased slowly as the sample size increased.

\begin{figure}[t]
\centering
\includegraphics[width=1\linewidth]{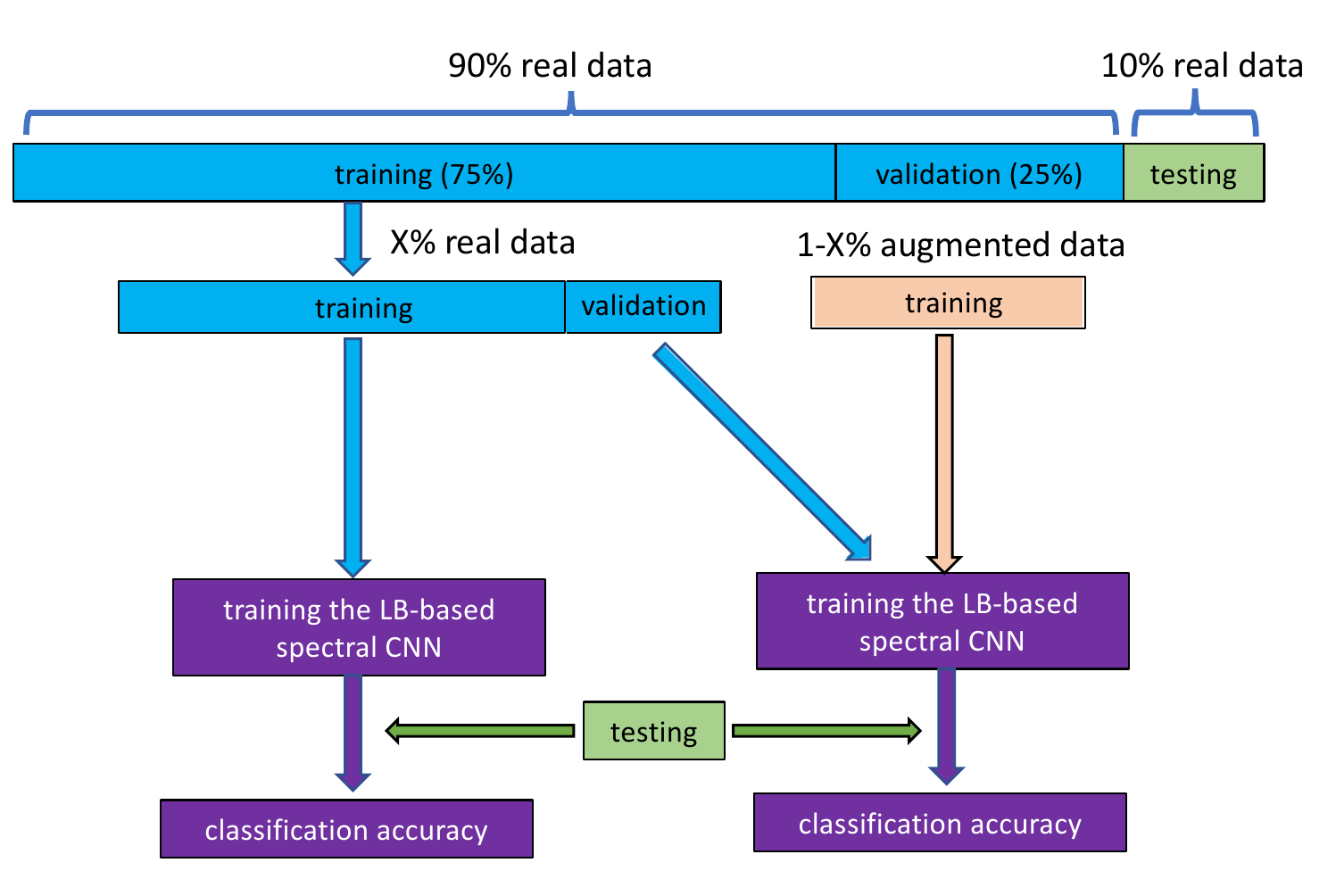}
\caption{Real and augmented data used in the LB-based spectral CNN. $X\%$ indicates that the percentage of the training set 
%and validation sets 
is real data and $1-X\%$ are augmented data. }
\label{fig:data_select}
\end{figure}

To demonstrate the use of the two augmented data in classification,  we fixed the total sample size as 3000 ($n=2000, \  m=1000$). Among the 3000 samples,  2025, 675, and 300 samples were respectively used as the training, validation, and testing samples.
%\footnote{This does not make any sense. For each 3000, there must be number for testing,t raining, validation. Same for 2025. So there should be 9 numbers here. So no idea what 300, 2025 and 675 refers to.}
 As illustrated in Fig. \ref{fig:data_select}, when only a smaller fraction of the simulated data in the training, denoted as $X\%$, was available, we applied the augmentation methods to add $1-X\%$ augmented data to the training set. For instance, if $X=10$, we only used $203$ of the training samples and employed LB-eigDA or C-pDA to generate $1822$ augmented data as additional training samples. The augmentation was employed separately for the two groups. The validation (675 samples) and testing (300 samples) sets remained the same. The classification accuracy was respectively $95.5\%$ for LB-eigDA and $92.5\%$ for C-pDA. Without the augmented data, the classification accuracy was $80.3\%$, more than $10\%$ lower than that obtained using the data augmented by LB-eigDA and C-pDA. 
Fig.~\ref{fig:simulation2}  (b) shows that LB-eigDA and C-pDA improved the classification accuracy when compared to that without augmented data.
%\footnote{compared to what? "that" means what. Writing in this paragraph can be significantly improved.} 
 Moreover, the LB-eigDA method performed in general better than the C-pDA method. This is mainly because the C-pDA method employs the polynomial approximation of the LB spectral filters.

\section{Results}
\label{sec:results}

We used MRI data from ADNI. We first illustrate the similarity of augmented data by  LB-eigDA and C-pDA to real MRI data. We then compare the computational cost of the LB-eigDA and C-pDA approaches. Finally, we show the use of C-pDA in the LB-based spectral CNN to improve the classification accuracy of Alzheimer's patients.

\subsection{MRI data acquisition and preprocessing}
We used ADNI-2 cohort (\protect\url{adni.loni.ucla.edu}) acquired from participants aging from 55 to 90 using either 1.5 or 3T scanners. For the typical 1.5T acquisition, repetition time (TR) $=2400$ ms, minimum full echo time (TE) and inversion time (TI)$ =1000$ ms, flip angle$ =8^\circ$, field-of-view (FOV)$ =240\times240$ mm$^2$, acquisition matrix$ =256\times256\times170$ in the x-, y-, and z-dimensions, yielding a voxel size of $1.25\times1.25\times1.2$ mm$^3$. For the 3T scans, TR$=2300$ ms, minimum full TE and TI $=900$ ms, flip angle$=8^\circ$, FOV$=260\times260$ mm$^2$, acquisition matrix $=256\times256\times170$, yielding a voxel size of $1.0\times1.0\times1.2$ mm$^3$.

We utilized the structural T1-weighted MRI from the ADNI-2 dataset. The number of visits of each subject varied from 1 to 7 (i.e., baseline, 3-, 6-, 12-, 24-, 36-, and 48-month), and at each visit, the subjects were diagnosed with one of the four clinical statuses based on the criteria in the ADNI protocol  (\protect\url{adni.loni.ucla.edu}): healthy control (HC), early mild cognitive impairment (MCI), late MCI, and Alzheimer’s disease (AD). In this study, we illustrated the use of the augmentation methods via the HC/AD classification since it has been well studied using T1-weighted image data (e.g., 
\cite{cuingnet2011automatic,liu2013locally,hosseini2016alzheimer,korolev2017residual,liu2018landmark,islam2018brain,basaia2019automated,wee2019cortical}). 
Hence, this study involved 643 subjects with HC or AD scans (392 subjects had HC scans; 253 subjects had AD scans). There were 8 subjects who fell into both groups due to the conversion from HC to AD. Tables \ref{tab:adni2} lists the demographic information of the  ADNI-2 cohort.

The T1-weighted images were segmented using FreeSurfer (version 5.3.0) \cite{fischl2002whole}. The white and pial cortical surfaces were generated at the boundary between white and gray matter and the boundary of gray matter and CSF, respectively. Cortical thickness was computed as the distance between the white and pial cortical surfaces. It represents the depth of the cortical ribbon. We represented cortical thickness on the mean surface, the average between the white and pial cortical surfaces.  We employed large deformation diffeomorphic metric mapping (LDDMM) \cite{zhong2010quantitative,du2011whole}  to align individual cortical surfaces to the atlas and transferred the cortical thickness of each subject to the atlas. The cortical atlas surface was represented as a triangulated mesh with 655,360 triangles and 327,684 vertices. At each surface vertex, a  spline regression implemented by piecewise step functions \cite{james2013introduction} was performed to regress out the effects of age and gender. The residuals from the regression were used in the below LB-based spectral CNN.
%\footnote{It is unclear if you used cortical thickness (that regress out age and gender) on deep-learning or if you did regression independently of deep learning. Be precise here.}

\begin{table}[th]
\caption{Demographic information of the ADNI-2 cohort with MRI scans.}
\label{tab:adni2}
\center{\setstretch{1.5}
\begin{tabular}{| l | l | l | }
\hline
 & HC & AD\\
\hline
the number of subjects$^\dagger$ & 400 &  261\\
\hline
the number of scans & 1122 & 587 \\
\hline
gender (female/male) &  607/515 &  254/333\\
\hline
age (years; mean$\pm$SD) &  75.3$\pm$6.8  & 75.3$\pm$7.7\\
\hline
\end{tabular}}\\\small{$^\dagger$ There are 8 subjects who fall into both the HC and AD groups due to the conversion from HC to AD. Abbreviations: HC, healthy controls; AD: Alzheimer's disease; SD, standard deviation.}
\end{table}

\subsection{LB-eigDA and C-pDA augmentation}
We extracted cortical thickness data from 500 ADNI brain MRI scans and then used them to generate augmented cortical thickness via LB-eigDA and C-pDA.

C-pDA requires determining the number of filters and the bandwidth of each filter. These parameters are dependent on the spectrum of real data and application specific. First, we analyzed the spectrum of cortical thickness data, which was predominantly in the low-frequency band. More filters with narrow bandwidth were needed in the low frequency, while fewer filters with wide bandwidth were needed in the high frequency. Second, the discrimination of cortical thickness between controls and AD patients lies in the low-frequency band. Hence, we empirically designed more filters in the low-frequency band based on the following procedure.

\begin{figure}[t]
\centering
\includegraphics[width=1\linewidth]{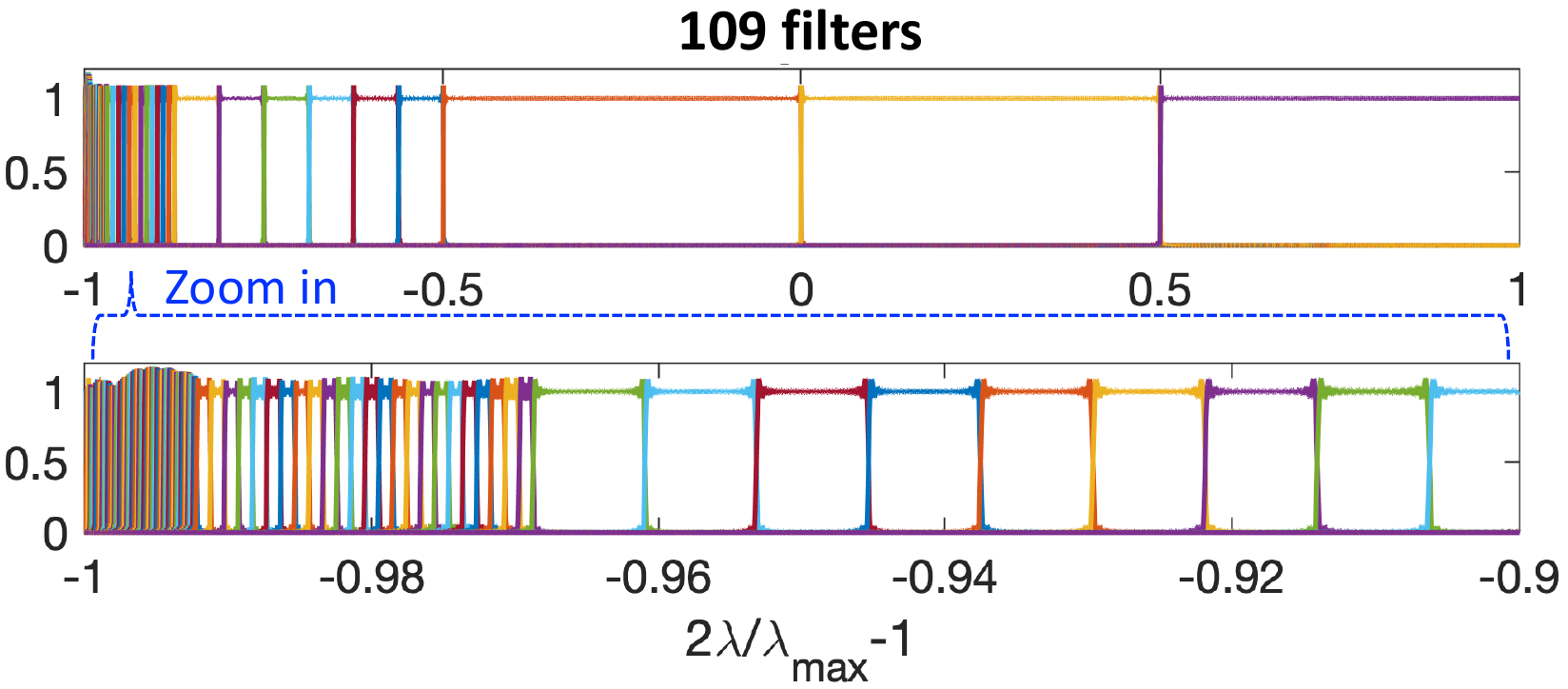}
\caption{A filter bank with 109 bandpass filters used in the C-pDA method.}
\label{fig:bp}
\end{figure}

Let $\lambda_{\max}$ be the maximum eigenvalue of the LB-operator of the cortical surface mesh.  We  divided the spectral range  of $[0,\frac{\lambda_{\max}}{4^{m-1}}]$ into $2^{m+1}$ equal-width  frequency bands, where $m$ is an integer between 1 and 5, and assigned a bandpass filter to each frequency band.  This procedure resulted in a total of 109 filters. 
%\footnote{What do you mean "one with a constant value"?????} 
Fig.~\ref{fig:bp} illustrates the filters used in this study. Moreover, the order of the Chebyshev polynomials needs to be determined  so that the transition of the filters is sharp. As illustrated in Fig.~\ref{fig:BW}, when $K=5000$, the approximation of the Chebyshev polynomials converges fast and has a small transition bandwidth. For the rest of this study,  we employed $K=5000$ for C-pDA. 

On the other hand, only one parameter, the number of LB eigenfunctions, is needed for LB-eigDA. This study used 5000 eigenfunctions for LB-eigDA, which covered the spectral range critical to the discrimination of controls and AD patients. %\footnote{C-pDA was explained in lengthy one paragraph. There is no explanation on LB-eigDA. Is it intentional or forgot to add material. It's just odd section}

\begin{figure*}[t]
\centering
\includegraphics[width=1\linewidth]{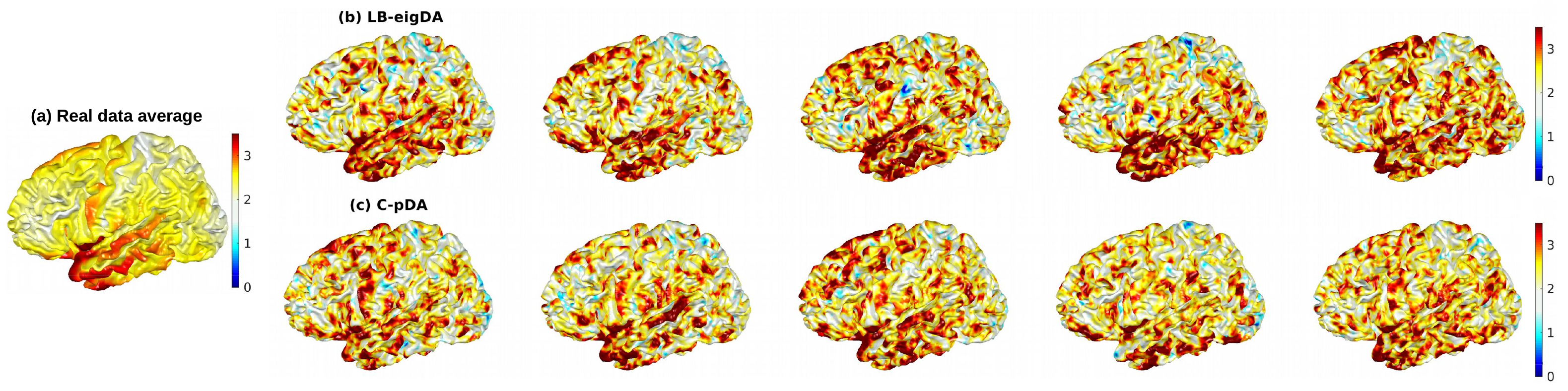}
\caption{Augmented cortical thickness. (a) Cortical thickness averaged over 500 real datasets; (b) five augmented thickness data via the LB-eigDA method; and
(c) five augmented thickness data via the C-pDA method.}
\label{fig:aug_example}
\end{figure*}

We employed LB-eigDA and C-pDA and generated 500 augmented cortical thickness data based on 500 randomly selected data from ADNI. Fig.~\ref{fig:aug_example}(a) illustrates cortical thickness averaged over the 500 real data.  Fig.~\ref{fig:aug_example}(b) and (c) show 5 augmented thickness data that were respectively generated by LB-eigDA and C-pDA. This figure suggests that the pattern of the augmented data from the two methods is similar to the averaged pattern observed in real data.

\begin{figure}[t]
\centering
\includegraphics[width=1\linewidth]{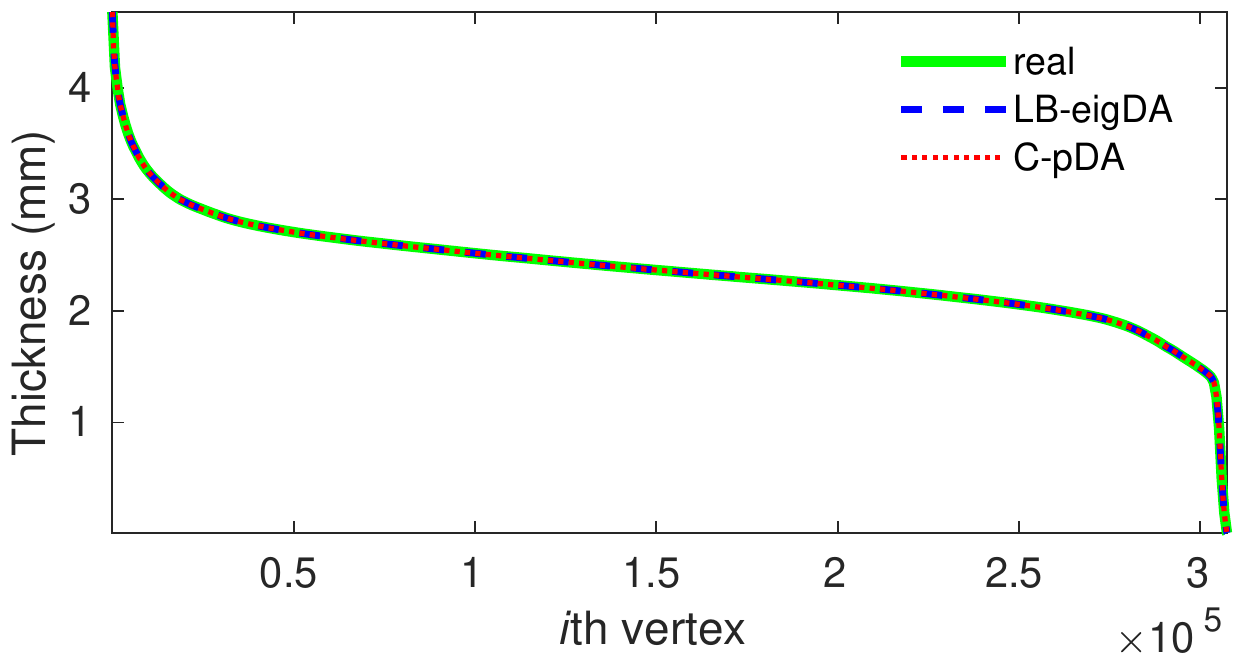}
\caption{Sorted thickness values at each vertex on the cortical surface mesh. The green solid, blue dashed, and red dotted lines represent the thickness value at a particular vertex averaged over the 500 real data, 500 augmented data via LB-eigDA, and 500 augmented data via C-pDA, respectively. For the purpose of visualization, the thickness averaged over 500 original data is sorted in a descend manner across all the vertices on the cortical surface mesh. The augmented data follow the sorted vertex index. }
\label{fig:aug_mean}
\end{figure}

\begin{figure}[t]
\centering
\includegraphics[width=1\linewidth]{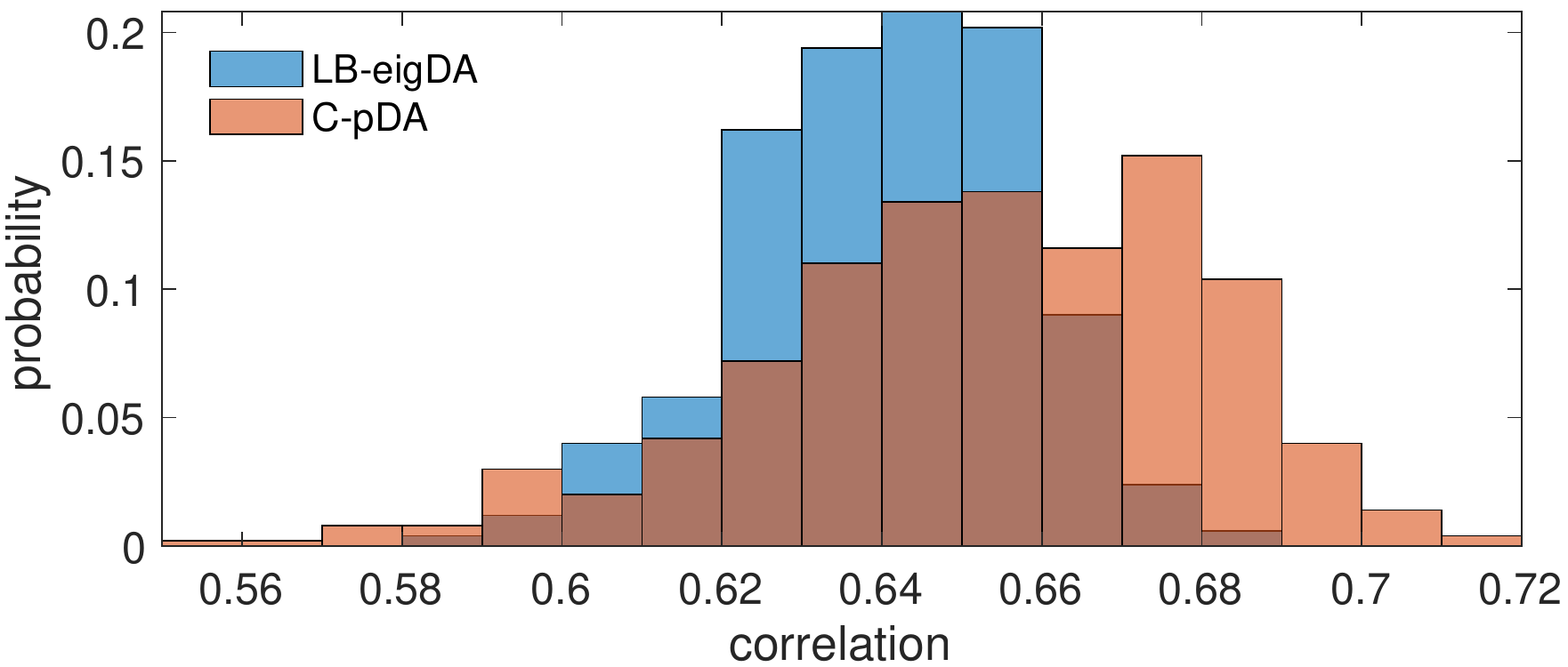}
\caption{The distribution of the correlation between the 500 augmented thickness data and the thickness averaged over 500 real data (blue bar for the LB-eigDA and orange bar for the C-pDA method. }
\label{fig:aug_corr}
\end{figure}

Moreover, Fig.~\ref{fig:aug_mean} shows the thickness averaged over the 500 real data (green solid line), the 500 LB-eigDA augmented data (blue dashed line), and the 500 C-pDA data (red dotted line), respectively. Both LB-eigDA and C-pDA preserved the mean of the real thickness data at each vertex of the cortical surface mesh. 
%\footnote{After 7 pages, you fail to explain why mean is preserved theoretically. Also you failed to explain why it even matter in application. Preserving mean has nothing do to with the classification accuracy! If it is related, demonstrate it.}
Empirically, the largest difference between the real and augmented data was smaller than $10^{-8}$ mm. Moreover, we computed Pearson's correlation of the averaged real data with the 500 augmented data. Fig.~\ref{fig:aug_corr} shows the distribution of these correlation values for the LB-eigDA and C-pDA methods. The correlation value of the LB-eigDA augmented thickness was in the range of $[0.58,\ 0.68]$ with mean and standard deviation of $0.64\pm0.02$, 
while the C-pDA augmented data showed the correlation in the range of $[0.53,\ 0.72]$ with mean and standard deviation of $0.65\pm0.03$. Overall, both the LB-eigDA and C-pDA methods can generate new data whose pattern is similar to that of real data.

\begin{figure}[t]
\centering
\includegraphics[width=1\linewidth]{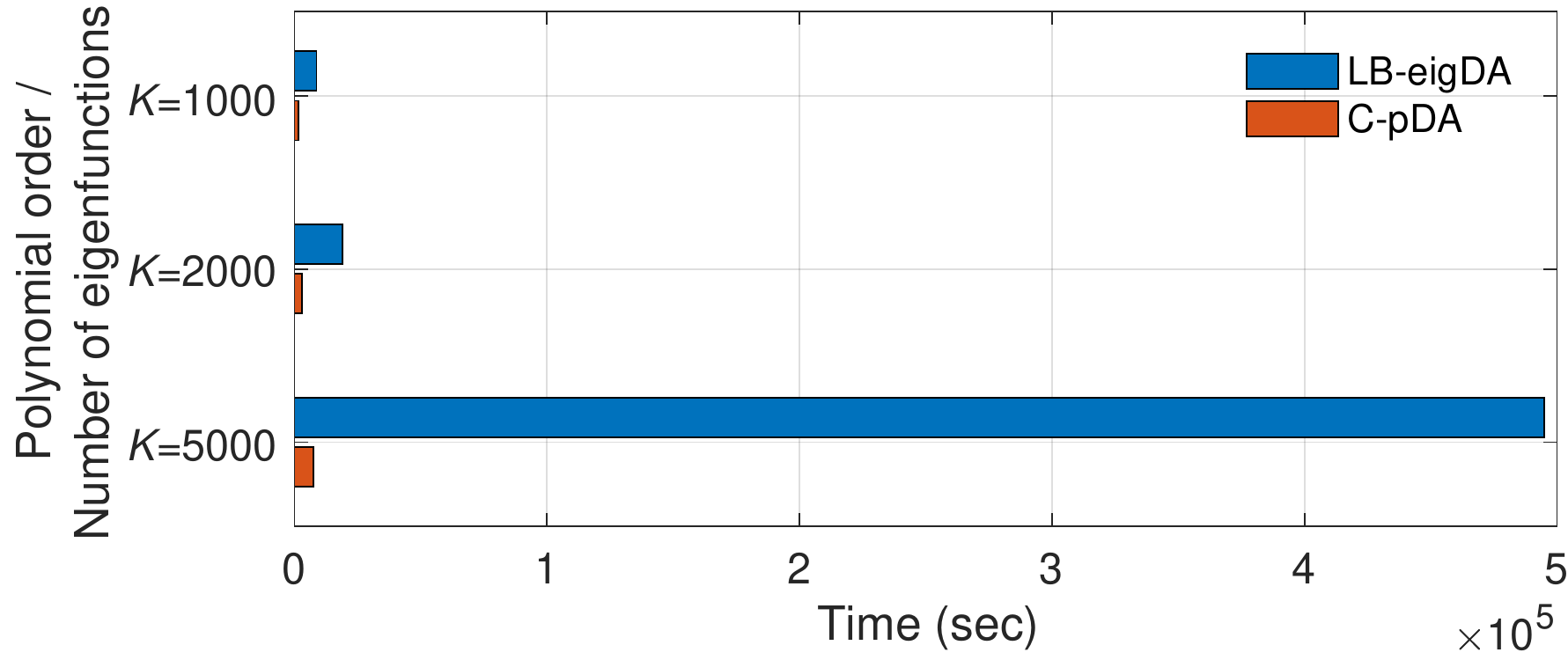}
\caption{Computational time of the LB-eigDA (blue) and C-pDA (red-orange) methods for generating 500 augmented thickness data from 500  randomly selected subjects in ADNI.}
\label{fig:eigaug_time}
\end{figure}

The LB-eigDA computational time was dependent on the number of the LB eigenfunctions, while the C-pDA computational time was related to the order of Chebyshev polynomials. Fig.~\ref{fig:eigaug_time} shows the LB-eigDA computational time  as a function of the number of the LB eigenfunctions and the C-pDA computational time as the order of Chebyshev polynomials, $K$. This figure suggests that more LB eigenfunctions used in LB-eigDA allow the augmentation over a wider spectrum but require a high computational cost when the cortical surface mesh is large (the cortical surface mesh with 327,684 vertices). The LB-eigDA computational cost was exponentially increased as the number of the LB eigenfunctions increased. In contrast, the C-pDA computational time was approximately a linear function of the order of Chebyshev polynomials. 
Compared to C-pDA , LB-eigDA was 70 times slower when  $K=5000$.

\subsection{Does classification improve by data augmentation?}\label{subsec:aug_improve}

%\begin{figure}[t]
%\centering
%\includegraphics[width=1\linewidth]{figs/data_select_v2.pdf}
%\caption{Real and augmented data used in the LB-based spectral CNN. $X\%$ indicates that the percentage of the training set 
%%and validation sets 
%is real data and $1-X\%$ are augmented data via C-pDA. }
%\label{fig:data_select}
%\end{figure}

We illustrate the use of the C-pDA method to classify healthy controls (HC) and AD patients based on the cortical thickness of the ADNI dataset. Again, we employed the LB-based spectral CNN with the architecture similar to that in Fig.~\ref{fig:architecture}, but used five convolutional layers. Each layer involved 8, 16, 32, 64, and 128 filters, respectively. The initial learning rate was $10^{-3}$, and the learning rate decay was $0.05$ for every $20$ epochs. In this experiment, the total sample from the ADNI dataset was $1709$ (HC: $ n=1122$; AD: $n=587$). Ten-fold cross-validation was adopted. One fold of real data was left out for testing. The remaining nine folds of data were further separated into training ($75\%$) and validation ($25\%$) sets. When the MRI datasets were separated into the training, validation, and testing sets, we considered subjects instead of MRI scans so that the scans from the same subjects were in the same set to avoid potential data over leakage.

\begin{figure}[t]
\centering
\includegraphics[width=1\linewidth,clip=true]{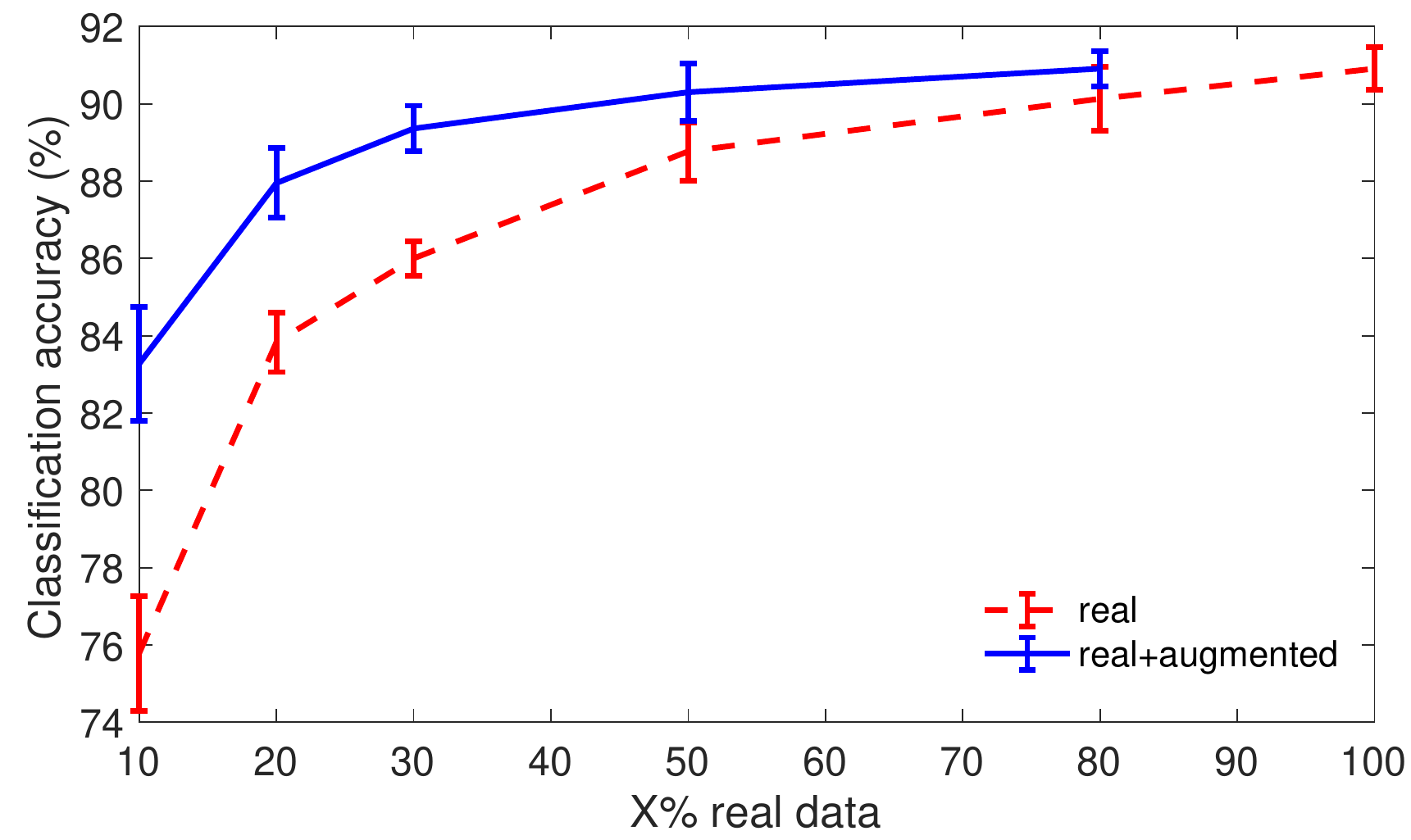}
\caption{Classification accuracy. The red dashed line shows the classification accuracy when only $X\%$ of the real training samples were used in the training of the LB-based spectral CNN. The blue solid line shows the classification accuracy when only $X\%$ of the real raining set and $1-X\%$ of the augmented data were used in the training of the LB-based spectral CNN, where the augmented data were generated by the C-pDA method. }
\label{fig:aug_acc}
\end{figure}

The HC/AD classification accuracy based on the real ADNI data and the LB-based spectral CNN was $90.9\pm0.6\%$.  However, when only a smaller set of the real data was available ($X\%$ of the training set), that is, the training sample size was reduced, the classification accuracy dropped as illustrated by the red dashed line in Fig.\ref{fig:aug_acc}. When only $10\%$ of the real data was available, the classification accuracy was $75.8\%$ and decreased $15\%$ compared to that using the full ADNI data.

We previously showed that both C-pDA and LB-eigDA have the same results but C-pDA was more computationally efficient than LB-eigDA. Thus, the following experiments only used C-pDA with 109 filters and the Chebyshev polynomials of order $K=5000$. As illustrated in Fig. \ref{fig:data_select}, the training samples contained $X\%$ of real ADNI data and $1-X\%$ 
augmented data, where $X=10,20, \cdots, 80$. We  added $1-X\%$ augmented data using C-pDA in the LB-based spectral CNN and computed the network performance using the testing real data. The augmentation was done separately for the HC and AD groups. For instance,  when $90\%$ of the training samples were augmented data and $10\%$ of  the training samples were real data, the classification accuracy was $83.3\%$ and improved by $7.5\%$. Fig.\ref{fig:aug_acc} shows that C-pDA can increase the sample size and improve the HC/AD classification accuracy.

\section{Discussion}

This study introduces the LB-eigDA and C-pDA methods to generate augmented data on surfaces. Using the simulation with the ground truth label, we demonstrate that both methods  improve the performance of graph-CNN. In particular,  LB-eigDA has the  potential to outperform  C-pDA method since C-pDA approximates the LB spectral filters using Chebyshev polynomials. Nevertheless, when the mesh becomes large, LB-eigDA is computationally intensive while  C-pDA is computationally efficient. C-pDA generates augmented thickness data and improves the AD classification accuracy in a real clinical application. 

To our best knowledge, this study provides the first unbiased oversampling
 approaches for data augmentation on surfaces. These methods have a  great potential to open new research areas in graph CNN in conjunction with generative adversarial networks (GANs). In particular, the formulation of the C-pDA method is consistent with that the LB-based spectral CNN  \cite{Defferrard2016,wee2019cortical}, which is feasible to adapt the C-pDA and graph network to the GAN framework. Further investigation will be needed.

\section*{Acknowledgements}
This research/project is supported by the National Science Foundation MDS-2010778, National Institute of Health R01 EB022856, EB02875, and National Research Foundation, Singapore under its AI Singapore Programme (AISG Award No: AISG-GC-2019-002). Additional funding is provided by the Singapore Ministry of Education (Academic research fund Tier 1; NUHSRO/2017/052/T1-SRP-Partnership/01), NUS Institute of Data Science. This research was also supported by the A*STAR Computational Resource Centre through the use of its high-performance computing facilities.

\bibliographystyle{IEEEtran}
\bibliography{DataAug_09212020}

\end{document}